\documentclass[10pt,twocolumn,letterpaper]{article}

\usepackage{cvpr}              
\usepackage[accsupp]{axessibility}  

\definecolor{cvprblue}{rgb}{0.21,0.49,0.74}
\usepackage[pagebackref,breaklinks,colorlinks,allcolors=cvprblue]{hyperref}

\def\paperID{19} 
\def\confName{CVPR}
\def\confYear{2026}

\title{Revisiting Radar Perception With Spectral Point Clouds}

\author{
Hamza Alsharif$^{1}$ \quad Jing Gu$^{1}$ \quad Pavol Jancura$^{1}$ \quad Satish Ravindran$^{2}$ \quad Gijs Dubbelman$^{1}$ \\[4pt]
$^{1}$Eindhoven University of Technology \quad $^{2}$NXP Semiconductors \\[2pt]
{\small \texttt{\{h.s.f.alsharif, j.gu1, p.jancura, g.dubbelman\}@tue.nl}} \quad
{\small \texttt{satish.ravindran@nxp.com}}
}

\begin{document}
\maketitle
\begin{abstract}
Radar perception models are trained with different inputs, from range-Doppler spectra to sparse point clouds.
Dense spectra are assumed to outperform sparse point clouds, yet they can vary considerably across sensors and configurations, which hinders transfer.
In this paper, we provide alternatives for
incorporating spectral information into radar point clouds and show that, point clouds need not underperform compared to spectra.
We introduce the spectral point cloud paradigm, where point clouds are treated as sparse, compressed representations of the radar spectra, and argue that, when enriched with spectral information, they serve as strong candidates for a unified input representation that is more robust against sensor-specific differences.
We develop an experimental framework
that compares spectral point cloud (PC) models at varying densities against a dense range-Doppler (RD) benchmark, and report the density levels where the PC configurations meet the performance of the RD benchmark.
Furthermore, we experiment with two basic spectral enrichment approaches,
that inject additional target-relevant information into the point clouds.
Contrary to the common belief that the dense RD approach is
superior, we show that point clouds can do just as well, and can surpass the RD benchmark when enrichment is applied.
Spectral point clouds can therefore serve as strong candidates for unified radar perception, paving the way for future radar foundation models.~\footnote{Project page available at: \url{https://www.tue-mps.org/Spectral-Point-Clouds-Radar/}.}
\end{abstract}    
\section{Introduction}
\label{sec:intro}

\begin{figure}[t]
    \centering
    \includegraphics[width=\linewidth]{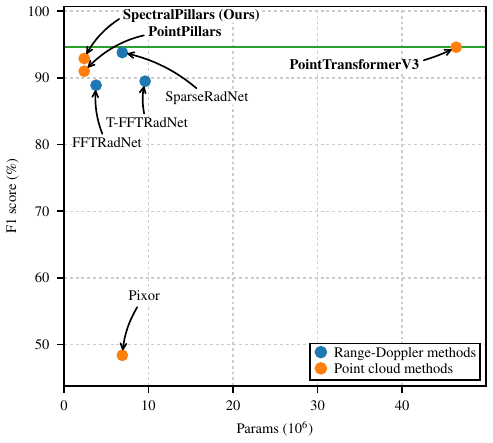}
    \caption{\textbf{Comparing radar perception approaches} on RADIal~\cite{RADIal2022} object detection.
    Point cloud models can compete with dense RD spectra models, and \textit{spectral enrichment} pushes their performance further.
    }
   \label{fig:sota_eye_catcher}
\end{figure}

Automotive imaging radars present new opportunities for improving the safety of autonomous vehicles.
Their reliability under adverse weather and poor lighting, together with the unique direct velocity measurements they provide, offer advantages that are more challenging to obtain with cameras or LiDAR.
As the angular resolution and consequent imaging capabilities of automotive radars improve, adopting radar for perception becomes an attractive prospect.

Most radar deep learning approaches use one of two radar input types: dense spectral tensors (range-Doppler, range-azimuth, range-azimuth-Doppler, etc.)~\cite{sparseradnet2024, carrada2021, erasenet_2023}, or sparse point clouds~\cite{VoD2022, mvfan_yan2023, smurf2024} derived from those spectra.
Given this diversity of input representations, there is currently no established understanding of which radar input representation is best suited for deep learning tasks.
Furthermore, although practitioners recognize the inherent sparsity of radar data, the question of how much input information radar models actually require for semantic tasks remains largely unanswered.

Most imaging radars expose only processed point clouds, with only a few exceptions providing access to dense spectral tensors.
Therefore, application developers have little control over which spectral features are preserved and how much information is retained for perception.
Furthermore, even when spectra are available, they vary substantially across sensors due to waveform design, antenna configuration, and transmission patterns.
This heterogeneity limits transfer and motivates a more standardized radar input representation, especially for large-scale pretraining and foundation models.

Training with RD spectra was initially shown in~\cite{RADIal2022} where the authors trained a convolutional architecture, FFTRadNet, on a detection and segmentation task.
They compared their RD approach to a point cloud (PC) baseline using Pixor~\cite{pixor2018}, and showed that the RD model significantly outperforms the PC model.
Additionally, they claim that point cloud training is more expensive due to the cost of the MIMO processing and associated angle-of-arrival (AoA) estimation steps, in addition to the voxelization of points onto a 3D grid. However, automotive radars typically offload much of the point cloud signal processing (including AoA estimation) to dedicated accelerators, reducing the burden on the main compute unit where the perception model runs.
Moreover, many point cloud methods operate on a 2D bird's eye view (BEV) grid in pillar-based approaches~\cite{pointpillars2019}, or directly in the space of the point sets~\cite{PTv3_2024}, rather than on a full 3D voxel grid.
Operating within a 3D grid would unnecessarily blow up model complexity, resulting in an inefficient point cloud model.

Since radar point clouds are derived from the same underlying spectra, they can preserve much of the target-relevant spectral information, while avoiding the cost of processing the full RD map.
We therefore hypothesize that, under matched data conditions, i.e., when both representations are constructed from the same set of detected spectrum peaks, a well-designed point cloud detector can match RD models, and that enriching points with target-relevant spectral context can allow it to surpass RD pipelines that must infer such context implicitly. To test this hypothesis, we propose the spectral point cloud paradigm, in which radar point clouds are treated explicitly as compressed representations of the underlying spectra and are systematically enriched with target-relevant information from them.
In this view, spectral point clouds provide a more standardized basis for perception models across sensors, while still retaining the spectral detail needed to exploit radar's full potential in neural networks.

To explore this paradigm, we develop an experimental setup based on RADIal~\cite{RADIal2022} that assesses the performance of point cloud models against a dense RD benchmark, across varying data density.
We adopt FFTRadNet~\cite{RADIal2022} and PointPillars~\cite{pointpillars2019} for the RD and PC pipelines respectively.
Based on our experimental results, our contributions are as follows:

\begin{itemize}
    \item We demonstrate that, radar point cloud models need not underperform dense RD models (\cref{fig:sota_eye_catcher}), provided that radar-specific cues such as Doppler and angular information are made explicit in the point cloud.

    \item We propose two complementary enrichment strategies, RD neighborhood expansion and angle spectrum descriptors, and show that they allow point clouds to achieve the RD benchmark performance with less data.

    \item We show that combined enrichment, our \textit{SpectralPillars} approach, enables point clouds to consistently surpass sparse RD baselines from $3.6\%$ density onward, motivating the spectral point cloud paradigm as a robust pathway to standardized radar inputs for future foundation models.
\end{itemize}
\section{Related Work}
\label{sec:related_work}

\subsection{Automotive Radar Designs and Datasets}

\begin{figure}[t]
    \centering
    \includegraphics[width=\linewidth]{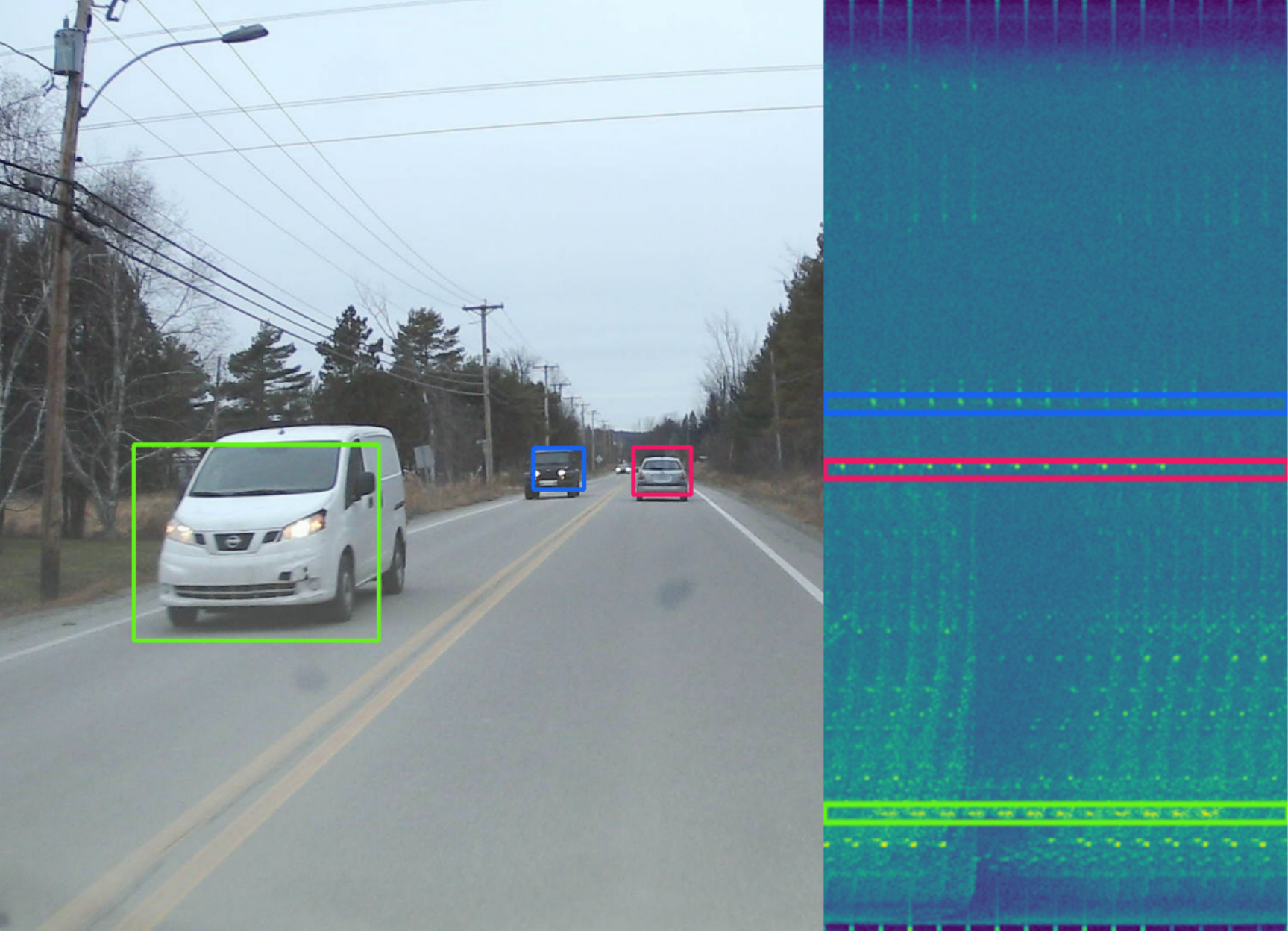}
    \caption{\textbf{RD spectrum visualization} with corresponding ground truth bounding boxes.
    The RD frame is incoherently integrated across Rx and visualized in log scale.
    The sensor uses simultaneous-Tx phase coding that interleaves per-Tx signatures along the Doppler axis (DDMA), yielding repeating patterns.
}
   \label{fig:sample_img_FFT}
\end{figure}

The availability of radar datasets is considerably lower than that of camera and LiDAR. 
Most HD imaging radar datasets provide only vendor-processed point clouds~\cite{tj4dradset2022, VoD2022, dualradar2025, mantruckscenes2024}, making any comparison against corresponding spectra infeasible.
Many other datasets provide raw radar data, but from a low-resolution radar~\cite{carrada2021, raddet2021, cruw2021, radical2021, Huang_2025_ICCV}, and are not representative of modern imaging radars.

Beyond dataset scarcity, a key barrier to unified radar perception is the tight coupling between radar design choices and the exported representations that learning methods are built around.
Datasets expose fundamentally different ``native" radar tensors, each embedding sensor-specific structure.
For example, RADIal~\cite{RADIal2022} provides HD per-receiver RD spectra, but the underlying radar employs a DDMA transmission scheme~\cite{Xu_beamspace_ddma2021}, which manifests as a repeating Tx-signature pattern along the Doppler axis (\cref{fig:sample_img_FFT}).
Methods trained directly on RD therefore implicitly learn in the presence of this spectral structure.
In contrast, K-Radar~\cite{kradar2022} releases fully FFT-processed range-azimuth-elevation-Doppler (RAED) tensors rather than per-receiver RD spectra.

Raw low-resolution radar datasets also differ in what level of the signal chain they expose and what processing is already baked into the tensors.
CRUW~\cite{cruw2021} releases radar data in pre-formed range-azimuth (RA) views, which fixes choices such as angular processing in the dataset representation and constrains access to phase-coherent information.
Similarly, CARRADA~\cite{carrada2021} organizes inputs as range-azimuth-Doppler (RAD) tensors, which again defines the learning input space in a way that is tightly linked to the sensor configuration and processing pipeline.

At the other end of the stack, many HD datasets expose only processed point clouds~\cite{tj4dradset2022, VoD2022, dualradar2025, mantruckscenes2024}, where proprietary filtering and signal processing decisions determine what information survives and in what form. 
This motivates releasing richer raw per-receiver spectra alongside point clouds, so that learning practitioners can redesign the representation for transfer and standardization.

Taken together, these factors mean that only a small subset of datasets enables controlled, matched comparisons between spectra and point clouds.
To the best of our knowledge, the only datasets that include raw spectrum data that comes from a HD imaging radar are RADIal~\cite{RADIal2022}, K-Radar~\cite{kradar2022}, and RaDelft~\cite{radelft2024}.
K-Radar does not provide per-receiver RD spectra, and RaDelft is lacking in terms of high-quality supervision labels.
Therefore, we adopt RADIal for our experiments.

\subsection{Radar Point Cloud Perception}

PointPillars~\cite{pointpillars2019} aggregates points into vertical pillars, encodes them with PointNets~\cite{pointnet2017} in each pillar, and applies efficient 2D convolutions on a BEV feature map. View-of-Delft~\cite{VoD2022} adapts PointPillars to radar and shows that radar-specific cues such as Doppler and RCS can improve detection.
Other works address radar point cloud sparsity and irregularity by reweighting and augmenting features.
For example MVFAN~\cite{mvfan_yan2023} fuses BEV and cylindrical views with saliency-based feature reweighting.
Attention-based approaches such as RPFA-Net~\cite{rpfa_net2021} and RadarPillars~\cite{radarpillars2024} use intra-/inter-pillar self-attention to capture broader context.
SMURF~\cite{smurf2024} further targets noise and sparsity by augmenting pillar features with a kernel density estimation branch~\cite{kde_chen2017}.

\subsection{Radar Spectra Perception}

Radar spectrum-based perception is mostly studied on low-resolution radar datasets~\cite{carrada2021, raddet2021, cruw2021, cruw3d2024, scorp2020, radical2021}, with relatively few high-resolution benchmarks~\cite{kradar2022, RADIal2022, radelft2024}.
Most methods incorporate angular information via range-azimuth (RA) representations~\cite{carrada2021, rodnet2021, Dong2020, madani2022radatron}, or full range-azimuth-Doppler (RAD) cubes~\cite{erasenet_2023, raddet2021, transrad2025}, and some use multi-view designs that jointly process RA/RD/AD views~\cite{transradar2024, mvrae2024}.

On high-resolution radar, FFTRadNet~\cite{RADIal2022} introduced learning directly from RD spectra for detection and segmentation, producing a latent RA representation for downstream heads.
SparseRadNet~\cite{sparseradnet2024} builds on this by explicitly exploiting spectral sparsity via learned subsampling~\cite{dps2021} and a sparse backbone.

Recent work also explores scaling with raw radar data.
In~\cite{Huang_2025_ICCV}, the authors introduce GRT, a transformer foundation model trained on a large low-resolution dataset using dense RAED tensors, and show that high-resolution imaging capabilities are achievable with large-scale pretraining.
While this highlights the promise of radar pretraining, operating on dense spectral tensors ties the input space closely to sensor configuration, motivating representations that are more suited to standardization across sensors.
\section{Methodology}

\begin{figure*}[t]
    \centering
    \includegraphics[width=\linewidth]{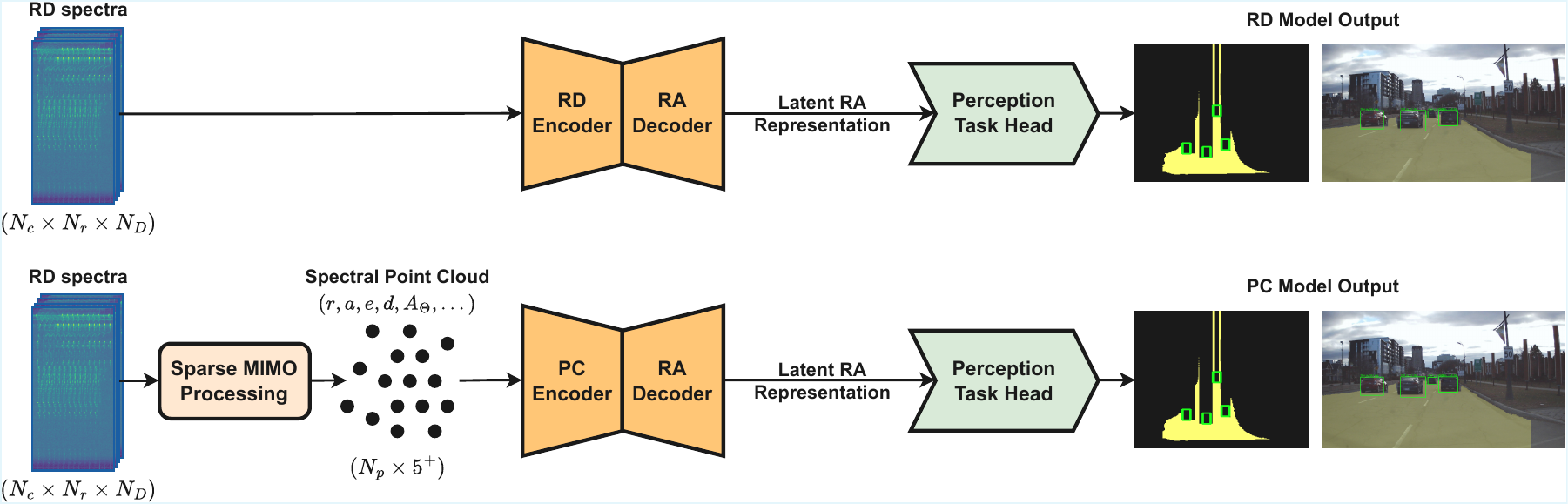}
    \caption{\textbf{RD and PC training pipelines}.
    The radar network can be trained with either RD spectra frames or spectral point clouds.
    In the RD pipeline, an implicit learnable RD to RA transformation is applied.
    In the PC pipeline, structure is imposed by a projection of latent features onto a BEV grid.
    }
   \label{fig:experiment_flows}
\end{figure*}

\cref{fig:experiment_flows} illustrates the training pipelines for the RD and PC detectors.
The radar encoder-decoder network is trained with either RD spectra or spectral point clouds derived from those same spectra.
In our experiments we instantiate the RD branch with FFTRadNet~\cite{RADIal2022}, which implicitly learns an RD-to-RA transformation, and the PC branch with PointPillars~\cite{pointpillars2019}, which imposes structure by projecting point features onto a BEV grid before RA decoding.
We use a common RA-aligned detection head to isolate the effect of the input representation.

\subsection{Model Architectures}

\paragraph{FFTRadNet}

The RD branch adopts FFTRadNet~\cite{RADIal2022} which consists of a MIMO pre-encoder, an FPN backbone, and a range-angle (RA) decoder.
Given a dense multi-channel RD tensor, the MIMO pre-encoder extracts low-level signal features, and the FPN backbone produces multi-scale feature maps.
To form an RA representation, the decoder folds Doppler information into the feature dimension (via a Doppler-channel axis swap) and applies convolutions that learn an implicit RD-to-RA mapping.
The multi-scale maps are then upsampled and fused to yield a latent RA grid, which is consumed by the task head (detection, and segmentation when used).

\paragraph{PointPillars}

The PC branch starts from an unstructured set of points and imposes structure via pillarization in PointPillars~\cite{pointpillars2019}.
The network partitions points into vertical pillars on a bird's eye view (BEV) grid defined in polar radar coordinates $(r, a)$, applies PointNets~\cite{pointnet2017} within pillars to get per-pillar embeddings, and scatters them to a dense BEV canvas.
A multi-scale BEV convolutional backbone then extracts features from this canvas yielding dense multi-scale representations analogous to the RD branch.
Since the BEV features are already RA-aligned and carry explicit angle information, the decoder only upsamples the multi-scale features to a common resolution before fusing them by concatenation.
In our implementation, we modify PointPillars by reducing the channel widths in the BEV backbone while slightly increasing its depth, resulting in a more lightweight architecture with fewer parameters.
Additionally, we swap out the activation functions from ReLU to GELU~\cite{hendrycks2016gaussian}, as we found empirically that this yields higher detection performance.
We use a BEV pillar grid with a spatial size of $256 \times 448$ in range-azimuth, which is downsampled to $128 \times 224$ to match the resolution of the output detection map.

\begin{figure}[t]
    \centering
    \includegraphics[width=\linewidth]{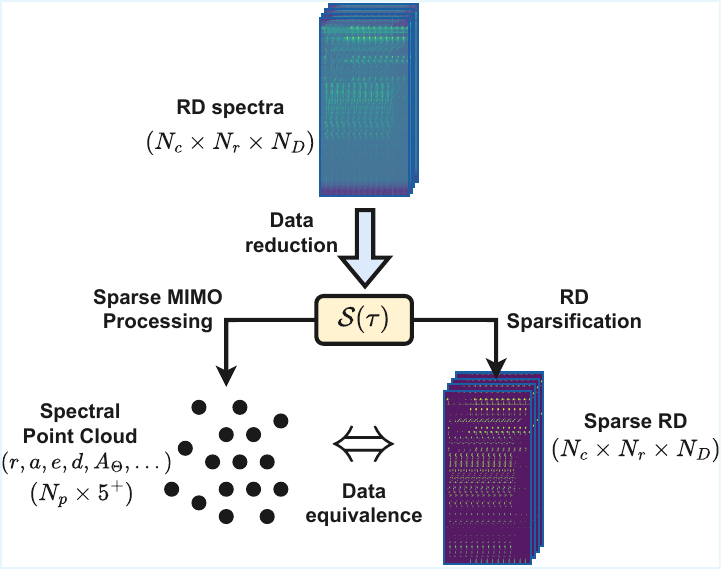}
    \caption{\textbf{Spectral point cloud data equivalence to RD}.
        A spectral point cloud is equivalent in data to a sparse RD frame with cells from peak set $\mathcal{S}(\tau)$, after CFAR-based data reduction.
    }
   \label{fig:representation_equivalence}
\end{figure}

\subsection{Spectral Point Cloud Generation}

Spectral point clouds treat radar point sets as \textit{compressed spectra}.
Each point is spatially located, but is \textit{spectrally grounded}: it is generated by summarizing a fixed subset of complex RD measurements associated with a detected peak (a virtual array snapshot).
By extracting this snapshot at each retained peak and applying sparse MIMO processing, we obtain angle spectrum attributes (AoA-derived angles and amplitudes) alongside the range and Doppler attributes.
This establishes a notion of \textit{data equivalence} between a point cloud and the underlying spectra: the point set is a compact encoding of a subset of RD cells.

We control retained spectral information through a shared peak index set $\mathcal{S}(\tau)$, obtained by CFAR-based reduction with threshold $\tau$.
The size of this set defines a \textit{data budget} for spectral point clouds and induces a natural density metric as the fraction of RD cells retained from the full grid.
Sweeping $\tau$ produces spectral point clouds across a range of budgets (densities), which we use to train configurations and compare against a dense RD benchmark that uses full information.
A sparse RD instance can also be formed from the same $\mathcal{S}(\tau)$, which we use only in the final representation comparison experiment.
\cref{fig:representation_equivalence} visualizes this correspondence.

\paragraph{CFAR as a Density Controller}
\label{sec:cfar_density_control}

We start from the complex per-receiver range-Doppler spectra $R \in \mathbb{C}^{N_c \times N_r \times N_D}$, indexed by receiver channel $c$, range $r$, and Doppler $d$.
The DDMA radar results in Tx-dependent replicas along the Doppler axis, yielding the repeating structure visible in~\cref{fig:sample_img_FFT}.
Since these repetitions are a radar configuration artifact, we consolidate them prior to applying CFAR.

We form an RD envelope $E \in \mathbb{R}^{N_r \times N_D^{*}}$ that is invariant to the transmit code pattern by incoherently summing power across Rx and then summing across Tx-replica bins, where $N_D^{*}$ is the length of the consolidated Doppler axis.
Running CFAR on $E$ yields a local noise estimate $\hat{\mu}(r, \ell)$. We perform the linear SNR test with threshold $\tau$:

\begin{equation}
  \mathcal{P}(\tau) \;=\; \Big\{\, (r,\; \ell) \;:\; \frac{E(r, \ell)}{\hat{\mu}(r, \ell)} \;>\; \tau\, \,\Big\}
\end{equation}

where $\ell \, \in \, \{0, \ldots , \; N_D^{*} - 1\}$ indexes the consolidated Doppler axis.

Each consolidated peak $(r, \; \ell)$ deterministically maps back to a fixed set of Doppler bins on the full axis $\mathcal{T}(r, \; \ell) \subset \{0, \ldots, N_D - 1\}$ where $|\mathcal{T}(r, \; \ell)| = M$ and $M$ equals the number of Tx signatures.
The corresponding set of peak locations on the full RD grid is:

\begin{equation}
    \mathcal{S}(\tau) = \bigcup_{(r, \; \ell) \in \mathcal{P}(\tau)} \Big\{ (r, \; d) : d \in \mathcal{T} (r, \; \ell) \Big\}
\end{equation}

The set $\mathcal{S}(\tau)$ thus defines a sparse set of ``information peaks" on the full RD grid.
We treat $\tau$ as a density controlling hyperparameter and report the resulting data density as:

\begin{equation}
    \text{Data density (\%)} \;=\;\frac{|\mathcal{S}(\tau)|}{N_r\,N_D}\times 100 \%
\end{equation}

\paragraph{Sparse MIMO Processing}

\label{sec:sparse_mimo_processing}

Let $N_{\rm virt}$ denote the number of virtual channels after stacking $N_c$ receivers across $M$ Tx-replica bins.
For each consolidated peak $j = (r_j, \ell_j) \in \mathcal{P}(\tau)$, we gather the Tx-replica bins on the full Doppler axis, $\mathcal{T}(r_j, \ell_j) \subset \{0, \ldots, N_D - 1\}$, and form a virtual array snapshot by stacking the complex channels:

\begin{equation}
    \mathbf{v}_j = \text{vec}(R[:, r_j, \mathcal{T}(r_j,\ell_j)]) \in \mathbb{C}^{N_{\rm virt}}
\end{equation}

We define a discrete azimuth-elevation grid with $N_\Theta$ directions, indexed by $k$, with $(a_k, e_k)$ denoting the azimuth and elevation of direction $k$.
Let $B \in \mathbb{C}^{N_\Theta \times N_{\rm virt}}$ be a pre-computed calibrated beamforming dictionary whose $k$-th row corresponds to direction $(a_k, e_k)$.
The angle spectrum for $j$ is given by:

\begin{equation}
    \mathbf{s}_j = |B \; \mathbf{v}_j|
\end{equation}

We take the dominant beam as the AoA estimate:

\begin{equation}
    k_j = \arg\max_k \mathbf{s}_j[k], \quad (a_j, e_j) = (a_{k_j}, e_{k_j})
\end{equation}

We also attach the associated angle spectrum amplitude at $k_j$ as an additional axis of information:

\begin{equation}
    A_{\Theta j} = \mathbf{s}_j[k_j]
\end{equation}

Using the consolidated Doppler index as the Doppler attribute $d_j = \ell_j$, the point coordinate $p_j$ for peak $j$ is then $(r_j, a_j, e_j, d_j, A_{\Theta j})$.

\subsection{Spectral Enrichment}

\label{sec:spectral_enrichment}

Enriching point clouds with additional spectral context can further improve deep learning performance.
In particular, we can exploit the local neighborhoods around each detected peak in $\mathcal{S}(\tau)$ to recover extended-target structure and spectral cues that are not captured by a single peak sample.
Here we introduce two such enrichment schemes for spectral point clouds: one that aggregates local neighborhoods in the RD domain into additional points, and another that aggregates angular neighborhoods in the angle-spectrum domain into additional per-point features.
Furthermore, we introduce the \textit{SpectralPillars} configuration where both enrichment schemes are used in conjunction with one another to get the most benefit out of spectral point clouds using PointPillars.
We compare the performance of SpectralPillars to sparse RD baselines in~\cref{sec:spec_pillars_sparse_rd}.

\paragraph{RD neighborhood enrichment}

Extended targets often occupy several neighboring RD bins rather than a single CFAR peak. To capture this local structure, we enlarge the CFAR detection map around each consolidated peak in $\mathcal{P}(\tau)$.
Let $n$ be an odd integer and define $h = \frac{n -1 }{2}$ as the neighborhood half-width.
For each consolidated peak $(r, \ell) \in \mathcal{P}(\tau)$ on the consolidated RD grid, we define its $n \times n$ neighborhood as:

\begin{equation}
    \mathcal{N}_n(r, \ell) = \{(r + \Delta r, \ell + \Delta \ell) : |\Delta r| \leq h, |\Delta \ell| \leq h \}
\end{equation}

$\mathcal{N}_n$ is truncated at the boundaries of the RD envelope.
The neighborhood augmented consolidated peak set is then:

\begin{equation}
    \mathcal{P}^*(\tau) = \bigcup_{(r, \ell) \in \mathcal{P}(\tau)} \mathcal{N}_n(r, \ell)
\end{equation}

$\mathcal{P}^*(\tau)$ is a superset of $\mathcal{P}(\tau)$, with up to $n^2 - 1$ neighboring bins added per CFAR detection in the interior of the grid.
We then apply the same subsequent processing steps in Section~\ref{sec:cfar_density_control} to obtain the corresponding neighborhood peak set $\mathcal{S}^*(\tau)$ which we generate points from.
Intuitively, RD neighborhood enrichment expands each CFAR detection by turning nearby RD cells into extra points, giving the spectral point cloud a denser footprint over extended targets.
In our experiments, we use $n = 3$ corresponding to a $3 \times 3$ neighborhood.

\paragraph{Angle spectrum enrichment}

In the standard point cloud processing chain (\cref{sec:sparse_mimo_processing}), for peak $j$ the azimuth and elevation bins are obtained from the
single strongest response in the angle spectrum vector $\textbf{s}_j$.
For a radar with $N_{\rm az}$ azimuth bins and $N_{\rm el}$ elevation bins, this means that $N_{\rm az} \times N_{\rm el} - 1$ angle spectrum bins are discarded for each peak, and the rich angular response pattern of the target is not used when forming the point cloud.
We introduce a simple method to compress this angular response into a fixed-length descriptor that is compatible with the standard ``maximum peak" AoA estimate.
Let $\textbf{s}_j$ be reshaped into an angle spectrum grid $S_j \in \mathbb{R}^{N_{\rm az} \times N_{\rm el}}$, with azimuth along one axis and elevation along the other.
We choose integers $n_{\rm az}$ and $n_{\rm el}$ that specify the number of angular sectors in azimuth and elevation respectively, and partition $S_j$ into a grid of $n_{\rm az} \times n_{\rm el}$ sectors.
For each sector indexed by $(u, v)$, we summarize its angular response by the maximum angle spectrum amplitude within that sector.
Let $\Omega_{u, v}$ be the set of indices $(a, e)$ that fall within sector $(u, v)$. We define the angular response of sector $(u, v)$ as:

\begin{equation}
    A^*_{\Theta j}(u, v) = \max_{(a, e) \in \Omega_{u, v}} S_j (a, e)
\end{equation}

This yields a pooled angle spectrum grid $A^*_{\Theta j} \in \mathbb{R}^{n_{\rm az} \times n_{\rm el}}$ where each entry stores the dominant response within one angular sector.
In practice, this operation corresponds to max-pooling $S_j$ over non-overlapping sectors, with kernel size and stride chosen such that the pooled output has spatial size $n_{\rm az} \times n_{\rm el}$.
We then flatten $A^*_{\Theta j}$ into a vector $\mathbf{a}_j \in \mathbb{R}^{n_{\rm az} n_{\rm el}}$ and attach $\mathbf{a}_j$ as additional feature dimensions of point $p_j$.
In all our experiments, we use angular pooling along azimuth only with $n_{\rm az} = 32$ and $n_{\rm el} = 1$.

\section{Experimental Setup}

Using selected CFAR thresholds $\tau$ spanning a range of data densities, we generate spectral point cloud datasets derived from the common peak set $\mathcal{S}(\tau)$ and train detectors across the sweep.
For all our experiments we use cell-averaging CFAR (CA-CFAR) with a $9\times9$ window with a $3\times 3$ guard window applied across the RD envelope $E \in \mathbb{R}^{N_r \times N_D^{*}}$, with $N_r = 512$, $N_D^{*} = 16$, and $N_D = 256$.
All sweep results are referenced to a dense RD benchmark trained on full RD spectra without data reduction.
When needed for representation-level comparison, we additionally generate matched sparse RD inputs from the same $\mathcal{S}(\tau)$.
Our main results focus on object detection, while additional segmentation results are reported in~\cref{tab:segmentation}.
To gauge the potential of stronger point-based architectures in the standard RADIal setting, we train PointTransformerV3~\cite{PTv3_2024} on spectral point clouds at the default RADIal density ($5.7\%$) and report it as a reference row in~\cref{tab:sota_comparison}.

All models are trained and evaluated on an NVIDIA H100 GPU using mixed precision for 50 epochs.
We train with AdamW~\cite{AdamW_loschilov2019} (weight decay = $0.01$) and a One-Cycle learning rate schedule~\cite{onecyclelr_smith2019} where the learning rate starts at $2.5 \times 10^{-4}$, rises to $2.5 \times 10^{-3}$, then cosine-anneals for the remainder of the training.
We use the same focal and smooth L1 combined loss configuration as in RADIal~\cite{RADIal2022}.
For each configuration, we train ten independent runs and report the mean test F1 score (at 0.5 IoU), where the test score is taken from the checkpoint that achieves the highest validation F1.
This is because the training runs are relatively noisy and have an overall standard deviation of $\pm 1.4\%$ F1.
When comparing against the state-of-the-art in~\cref{tab:sota_comparison}, we train for 100 epochs and follow the RADIal evaluation protocol, reporting results averaged over a sweep of confidence score thresholds.
For our density variation experiments, we assume a fixed-detector scenario and set the confidence threshold to $0.2$ for both spectral point clouds and RD inputs.
To summarize curve-level improvements across a density sweep, we compute the normalized area under the F1-density curve over a fixed interval using trapezoidal integration.
Dividing by the interval length yields the mean F1 over the sweep.
Differences between normalized area-under-the-curve values therefore correspond to average F1 gains across densities.
\section{Results}

\begin{figure}[t]
    \centering
    \includegraphics[width=\linewidth]{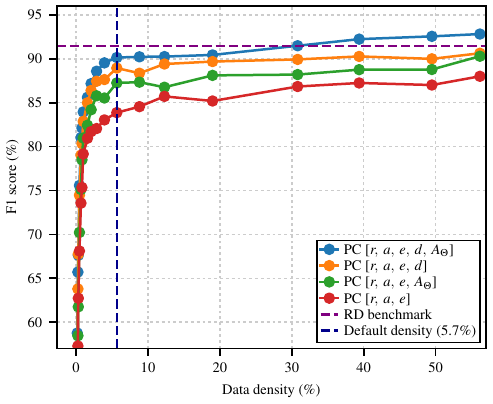}
    \caption{
        \textbf{PC F1 score trend, with axis ablations.}
        The angle amplitude $A_\Theta$ axis yields a $2.0$ point average improvement over spatial coordinates $[r, a, e]$, and the Doppler bin $d$ axis yields a $3.6$ point average improvement.
        Including both axes results in a $4.6$ point average improvement.
        At $30.8\%$ density, full-axes point clouds surpass the RD benchmark.
    }
   \label{fig:cloud_ablation_plot}
\end{figure}

\subsection{Spectral Point Clouds}

\cref{fig:cloud_ablation_plot} reports F1 against data density for point cloud variants obtained by adding Doppler $d$ and angle amplitude $A_\Theta$ attributes to the spatial coordinates $[r, a, e]$, alongside the dense RD benchmark.
Over the density sweep, adding $A_\Theta$ and $d$ yield average gains of $+2.0$ and $+3.6$ F1 over $[r, a, e]$, while combining both gives $+4.6$ F1, indicating Doppler as the dominant cue with $A_\Theta$ providing complementary benefit.
At the default density (as defined by RADIal) of $5.7\%$, the full spectral point cloud $[r, a, e, d, A_\Theta]$ is within $1.3$ F1 of the RD benchmark.
As density increases, it matches and then exceeds the benchmark at $30.8\%$, reaching $92.8$ F1 at $56.2\%$.
Notably, performance does not degrade with increased density, suggesting the detector benefits from retaining additional spectral information despite increased clutter.

Segmentation follows the same trend (\cref{tab:segmentation}): spectral point clouds match the RD benchmark from $1.6\%$ density onward and reach $81.1\%$ mIoU at $56.2\%$ density, which is a $+1.7$ mIoU over the benchmark.

\begin{table}[ht]
\centering
\caption{\textbf{Segmentation results} for various densities.}
\begin{tabular}{ccc}
\toprule
\textbf{Modality}     & \textbf{Density (\%)} & \textbf{mIoU (\%)} \\
\midrule
RD benchmark & 100          & 79.4 \\
\midrule
PC           & 0.5          & 78.3 \\
PC           & 1.6          & 79.6 \\
PC           & 5.7          & 80.6 \\
PC           & 19.0         & 80.7 \\
PC           & 30.8         & 81.0 \\
PC           & 56.2         & 81.1 \\
\bottomrule
\end{tabular}
\label{tab:segmentation}
\end{table}

\subsection{Spectral Enrichment}

\begin{figure}[t]
    \centering
    \includegraphics[width=\linewidth]{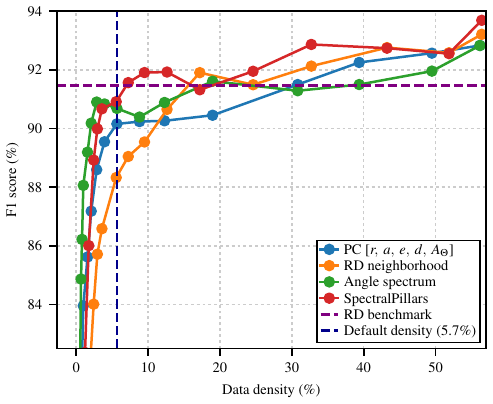}
    \caption{\textbf{PC enrichment scheme performance.}
    RD neighborhood enrichment reaches the RD benchmark at $17.2\%$ density but can slightly reduce F1 at very low densities, whereas angle spectrum enrichment reaches the benchmark at $19.0\%$ density without low-density degradation.
    SpectralPillars reaches the benchmark earliest, at $7.2\%$ density.
    }
   \label{fig:enriched_plot}
\end{figure}

\cref{fig:enriched_plot} compares the performance of enrichment strategies against that of the non-enriched baseline.
SpectralPillars is combined  RD neighborhood and angle spectrum enrichment.
RD neighborhood enrichment reaches the RD benchmark at $17.2\%$ density but slightly reduces F1 at the lowest densities, whereas angle spectrum enrichment reaches the benchmark at $19.0\%$ density while improving over the non-enriched baseline across most of the sweep.
This highlights the value of retaining angle spectrum information, which is typically discarded in standard point cloud pipelines.

The largest gains arise when both are applied together.
SpectralPillars reaches the RD benchmark at $7.2\%$ density, much earlier than any prior configuration.
At this operating point it also runs $89$ FPS faster than the RD detector (\cref{tab:fps_table}), making it simultaneously more accurate and more efficient in this regime.
Note that FPS reduces with increasing point cloud density, due to the increased cost of the point pillarization overhead.
Over the sweep, SpectralPillars improves over the non-enriched baseline by $+0.8$ average F1 and reaches $93.7\%$ F1 at $56.4\%$ density, a $+0.9$ F1 improvement over the corresponding non-enriched point.

This demonstrates how target-relevant spectral information can be exposed to the network using generic mechanisms: increasing point count (RD neighborhoods) and increasing per-point feature dimensionality (angle spectrum descriptors).
Because the schemes do not require reworking the backbone for a specific radar configuration, they motivate spectral point clouds as a practical basis for more standardized radar inputs and future radar foundation models.

\begin{table}[t]
\centering
\caption{\textbf{Model throughput} for various densities. FF: FFTRadNet, PP: PointPillars, SP: SpectralPillars. M.: Model, Ds.: Density.
}
\begin{tabular}{cccccc}
\toprule
\textbf{M.} & \textbf{Ds. (\%)} & \textbf{F1} & \textbf{SP. Ds. (\%)} & \textbf{SP. F1} & \textbf{FPS} \\
\midrule
FF             & 100                   & 91.5        & -                          & -                & 169        \\
\midrule
PP             & 0.7                   & 81.0        & 0.7                        & 74.3             & 279        \\
PP             & 1.6                   & 86        & 1.8                        & 86.0             & 276        \\
PP             & 5.7                   & 90.2        & 5.6                        & 90.9             & 264        \\
PP             & 8.8                   & 90.2        & 7.2                        & 91.6             & 258        \\
PP             & 19.0                  & 90.5        & 17.2                       & 91.3             & 232        \\
PP             & 30.8                  & 91.5        & 32.7                       & 92.9             & 207        \\
PP             & 56.2                  & 92.8        & 56.4                       & 93.7             & 192        \\
\bottomrule
\end{tabular}
\label{tab:fps_table}
\end{table}

\begin{table*}[ht]
\centering
\caption{
    \textbf{Comparison with state-of-the-art.}
    Bold text denotes the best result, and underlined denotes the second best.
    The last three rows are results that we have produced based on the default density ($5.7\%$), whereas the rest are taken from their respective papers.
    Pixor, considerably underperforms compared to RD results.
}
\begin{tabular}{cccccccc}
\toprule
\textbf{Model}                   & \textbf{Mode} & \textbf{Params ($10^6$)} & \textbf{AP  ($\%$)} & \textbf{AR ($\%$)} & \textbf{F1 ($\%$)} & \textbf{RE ($m$)} & \textbf{AE (\textdegree)} \\
\midrule
FFTRadNet~\cite{RADIal2022}     &       RD      &       \underline{3.8}     &       \textbf{96.8}        &       82.2        &       88.9        &       \underline{0.11}        &       0.17    \\
T-FFTRadNet~\cite{t_fftradnet2023}                      & RD            & 9.6                      & 89.5             & 89.6             & 89.5             & 0.15            & \underline{0.12}              \\
SparseRadNet~\cite{sparseradnet2024}                     & RD            & 6.9                      & 96.0             & 91.8             & \underline{93.8}             & 0.13            & \textbf{0.10}              \\
\midrule
Pixor~\cite{RADIal2022, pixor2018}                            & PC            & 6.9                      & 96.5             & 32.3             & 48.4             & 0.17            & 0.25              \\
PointPillars~\cite{pointpillars2019}                     & PC            & \textbf{2.4}                      & 88.7             & \textbf{93.5}             & 91.0             & 0.13            & 0.21              \\
SpectralPillars (Ours)                   & PC            & \textbf{2.4}                      & 94.3             & 91.5             & 92.9             & 0.12            & 0.22              \\
PointTransformerV3~\cite{PTv3_2024}                      & PC            & 46.4                             & \underline{96.6}              & \underline{92.7}             & \textbf{94.6}             & \textbf{0.10}         & \underline{0.12}          \\
\bottomrule
\end{tabular}

\label{tab:sota_comparison}
\end{table*}

\subsection{Comparing to Sparse RD}
\label{sec:spec_pillars_sparse_rd}

\begin{figure}[t]
    \centering
    \includegraphics[width=\linewidth]{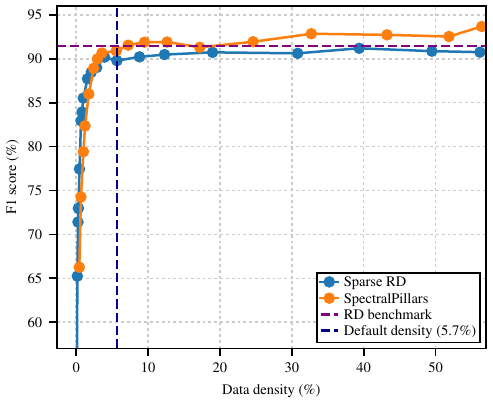}
    \caption{\textbf{SpectralPillars compared to sparse RD.}
    Average F1 gain over sparse RD is +1.3 points.
    }
   \label{fig:enriched_combined_vs_sparse_rd}
\end{figure}

\cref{fig:enriched_combined_vs_sparse_rd} compares SpectralPillars to sparse RD inputs trained with FFTRadNet under the same density sweep.
Sparse RD inputs are formed by masking the full RD spectra $R$ to retain only the peak cells indexed by $\mathcal{S}(\tau)$ (all other cells set to zero), ensuring matched data at each threshold $\tau$.
The SpectralPillars curve is above sparse RD from $3.6\%$ density onward, with a $+1.3$ average F1 gain over the sweep.
At $56.4\%$ density it reaches $93.7\%$ F1, exceeding the corresponding sparse RD run by $+2.9$ F1 and the dense RD benchmark by $+2.2$ F1.
This advantage is consistent with how enrichment exposes additional spectral information: RD neighborhood expansion adds local context around candidate peaks (capturing extended-target structure), while angle spectrum descriptors inject angular response information as per-point features.
Together, these cues let the SpectralPillars configuration close, and surpass, the sparse RD baseline at substantially lower densities.
A further structural difference may contribute: the point cloud branch provides explicit angular estimates (and angle spectrum descriptors), whereas the RD branch must recover angular information implicitly from RD spectra.
While stronger RD architectures could reduce this gap, our results suggest that making angular information explicit is an effective route to stronger sparse radar perception.

\subsection{Comparing to State-of-the-Art}

\cref{tab:sota_comparison} shows that the previously reported Pixor~\cite{pixor2018, RADIal2022} baseline performs far below RD methods, illustrating the gap from naive radar point cloud training.
Our radar-adapted PointPillars model closes most of this gap, reaching $91.0\%$ F1 with $2.4$M parameters.
With SpectralPillars, the F1 score reaches $92.9\%$ at the same model size, within $0.9$ F1 of the RD state-of-the-art while using $65\%$ fewer parameters.
Finally, PointTransformerV3~\cite{PTv3_2024} trained on spectral point clouds at the default density achieves the best overall detection performance ($94.6\%$ F1).
This upper bound indicates that, when point clouds are treated as compressed spectral representations and enriched with radar-specific cues, they can match or exceed RD-based detectors across model capacities.

\section{Limitations and Future Work}

Our study is conducted on RADIal, which provides high-resolution RD spectra and strong supervision, but its driving scenarios are relatively limited in diversity and complexity.
This restricts how broadly we can generalize the observed trends to more challenging conditions.
In future work, we aim to validate spectral point clouds and enrichment under more diverse conditions and larger data volumes, and assess robustness across radar configurations.
\section{Conclusion}

We presented spectral point clouds, an alternative radar perception representation that treats point clouds as spectral peaks and augments them with target-relevant spectral context.
This representation offers computational and memory benefits for resource-constrained edge devices, while providing a basis for standardization across radar designs and datasets.
We evaluated spectral point clouds against a dense RD benchmark, and showed that applying simple enrichment mechanisms, which yields our SpectralPillars configuration, closes the performance gap compared to the benchmark at lower densities and surpasses sparse RD baselines.
We hope this motivates broader study of matched representation comparisons and radar inputs that better support scaling to larger datasets and future foundation model settings.

\section*{Acknowledgments}

This work was partly funded by the NXP RAIDAR project, and the NWO LTP ROBUST project. Additionally, this work used the Dutch national e-infrastructure with the support of the SURF Cooperative using grant no. EINF-11179.
{
    \small
    \bibliographystyle{ieeenat_fullname}
    \bibliography{main}

@String(CVPR= {IEEE Conf. Comput. Vis. Pattern Recog.})

@String(ICCV= {Int. Conf. Comput. Vis.})

@String(ICPR = {Int. Conf. Pattern Recog.})

@String(CVPR  = {CVPR})

@String(ICCV  = {ICCV})

@String(ICPR  = {ICPR})

@inproceedings{sparseradnet2024,
    title={SparseRadNet: Sparse Perception Neural Network on Subsampled Radar Data},
    author={Wu, Jialong and Meuter, Mirko and Schoeler, Markus and Rottmann, Matthias},
    booktitle={European Conference on Computer Vision},
    pages={52--69},
    year={2024},
    organization={Springer}
}

@INPROCEEDINGS{carrada2021,
    author={Ouaknine, Arthur and Newson, Alasdair and Rebut, Julien and Tupin, Florence and Pérez, Patrick},
    booktitle={2020 25th International Conference on Pattern Recognition (ICPR)}, 
    title={CARRADA Dataset: Camera and Automotive Radar with Range- Angle- Doppler Annotations}, 
    year={2021},
    volume={},
    number={},
    pages={5068-5075},
    doi={10.1109/ICPR48806.2021.9413181}
}

@INPROCEEDINGS{erasenet_2023,
  author={Fang, Shihong and Zhu, Haoran and Bisla, Devansh and Choromanska, Anna and Ravindran, Satish and Ren, Dongyin and Wu, Ryan},
  booktitle={2023 IEEE International Conference on Robotics and Automation (ICRA)}, 
  title={ERASE-Net: Efficient Segmentation Networks for Automotive Radar Signals}, 
  year={2023},
  volume={},
  number={},
  pages={9331-9337},
  doi={10.1109/ICRA48891.2023.10160343}
}

@ARTICLE{VoD2022,
    author={Palffy, Andras and Pool, Ewoud and Baratam, Srimannarayana and Kooij, Julian F. P. and Gavrila, Dariu M.},
    journal={IEEE Robotics and Automation Letters}, 
    title={Multi-Class Road User Detection With 3+1D Radar in the View-of-Delft Dataset}, 
    year={2022},
    volume={7},
    number={2},
    pages={4961-4968},
    doi={10.1109/LRA.2022.3147324}
}

@inproceedings{mvfan_yan2023,
  title={Mvfan: Multi-view feature assisted network for 4d radar object detection},
  author={Yan, Qiao and Wang, Yihan},
  booktitle={International Conference on Neural Information Processing},
  pages={493--511},
  year={2023},
  organization={Springer}
}

@ARTICLE{smurf2024,
    author={Liu, Jianan and Zhao, Qiuchi and Xiong, Weiyi and Huang, Tao and Han, Qing-Long and Zhu, Bing},
    journal={IEEE Transactions on Intelligent Vehicles}, 
    title={SMURF: Spatial Multi-Representation Fusion for 3D Object Detection With 4D Imaging Radar}, 
    year={2024},
    volume={9},
    number={1},
    pages={799-812},
    doi={10.1109/TIV.2023.3322729}
}

@InProceedings{RADIal2022,
    author    = {Rebut, Julien and Ouaknine, Arthur and Malik, Waqas and P\'erez, Patrick},
    title     = {Raw High-Definition Radar for Multi-Task Learning},
    booktitle = {Proceedings of the IEEE/CVF Conference on Computer Vision and Pattern Recognition (CVPR)},
    month     = {June},
    year      = {2022},
    pages     = {17021-17030}
}

@InProceedings{pixor2018,
    author = {Yang, Bin and Luo, Wenjie and Urtasun, Raquel},
    title = {PIXOR: Real-Time 3D Object Detection From Point Clouds},
    booktitle = {Proceedings of the IEEE Conference on Computer Vision and Pattern Recognition (CVPR)},
    month = {June},
    year = {2018}
}

@InProceedings{pointpillars2019,
    author = {Lang, Alex H. and Vora, Sourabh and Caesar, Holger and Zhou, Lubing and Yang, Jiong and Beijbom, Oscar},
    title = {PointPillars: Fast Encoders for Object Detection From Point Clouds},
    booktitle = {Proceedings of the IEEE/CVF Conference on Computer Vision and Pattern Recognition (CVPR)},
    month = {June},
    year = {2019}
}

@InProceedings{PTv3_2024,
    author    = {Wu, Xiaoyang and Jiang, Li and Wang, Peng-Shuai and Liu, Zhijian and Liu, Xihui and Qiao, Yu and Ouyang, Wanli and He, Tong and Zhao, Hengshuang},
    title     = {Point Transformer V3: Simpler Faster Stronger},
    booktitle = {Proceedings of the IEEE/CVF Conference on Computer Vision and Pattern Recognition (CVPR)},
    month     = {June},
    year      = {2024},
    pages     = {4840-4851}
}

@INPROCEEDINGS{tj4dradset2022,
    author={Zheng, Lianqing and Ma, Zhixiong and Zhu, Xichan and Tan, Bin and Li, Sen and Long, Kai and Sun, Weiqi and Chen, Sihan and Zhang, Lu and Wan, Mengyue and Huang, Libo and Bai, Jie},
    booktitle={2022 IEEE 25th International Conference on Intelligent Transportation Systems (ITSC)}, 
    title={TJ4DRadSet: A 4D Radar Dataset for Autonomous Driving}, 
    year={2022},
    volume={},
    number={},
    pages={493-498},
    doi={10.1109/ITSC55140.2022.9922539}
}

@inproceedings{mantruckscenes2024,
    author = {Fent, Felix and Kuttenreich, Fabian and Ruch, Florian and Rizwin, Farija and Juergens, Stefan and Lechermann, Lorenz and Nissler, Christian and Perl, Andrea and Voll, Ulrich and Yan, Min and Lienkamp, Markus},
    booktitle = {Advances in Neural Information Processing Systems},
    editor = {A. Globerson and L. Mackey and D. Belgrave and A. Fan and U. Paquet and J. Tomczak and C. Zhang},
    pages = {62062--62082},
    publisher = {Curran Associates, Inc.},
    title = {MAN TruckScenes: A multimodal dataset for autonomous trucking in diverse conditions},
    volume = {37},
    year = {2024}
}

@article{dualradar2025,
    title={Dual radar: A multi-modal dataset with dual 4d radar for autononous driving},
    author={Zhang, Xinyu and Wang, Li and Chen, Jian and Fang, Cheng and Yang, Guangqi and Wang, Yichen and Yang, Lei and Song, Ziying and Liu, Lin and Zhang, Xiaofei and others},
    journal={Scientific data},
    volume={12},
    number={1},
    pages={439},
    year={2025},
    publisher={Nature Publishing Group UK London}
}

@INPROCEEDINGS{raddet2021,
    author={Zhang, Ao and Nowruzi, Farzan Erlik and Laganiere, Robert},
    booktitle={2021 18th Conference on Robots and Vision (CRV)}, 
    title={RADDet: Range-Azimuth-Doppler based Radar Object Detection for Dynamic Road Users}, 
    year={2021},
    volume={},
    number={},
    pages={95-102},
    doi={10.1109/CRV52889.2021.00021}
}

@InProceedings{cruw2021,
    author    = {Wang, Yizhou and Wang, Gaoang and Hsu, Hung-Min and Liu, Hui and Hwang, Jenq-Neng},
    title     = {Rethinking of Radar's Role: A Camera-Radar Dataset and Systematic Annotator via Coordinate Alignment},
    booktitle = {Proceedings of the IEEE/CVF Conference on Computer Vision and Pattern Recognition (CVPR) Workshops},
    month     = {June},
    year      = {2021},
    pages     = {2815-2824}
}

@ARTICLE{radical2021,
    author={Lim, Teck-Yian and Markowitz, Spencer A. and Do, Minh N.},
    journal={IEEE Journal of Selected Topics in Signal Processing}, 
    title={RaDICaL: A Synchronized FMCW Radar, Depth, IMU and RGB Camera Data Dataset With Low-Level FMCW Radar Signals}, 
    year={2021},
    volume={15},
    number={4},
    pages={941-953},
    doi={10.1109/JSTSP.2021.3061270}
}

@InProceedings{Huang_2025_ICCV,
    author    = {Huang, Tianshu and Prabhakara, Akarsh and Chen, Chuhan and Karhade, Jay and Ramanan, Deva and O'toole, Matthew and Rowe, Anthony},
    title     = {Towards Foundational Models for Single-Chip Radar},
    booktitle = {Proceedings of the IEEE/CVF International Conference on Computer Vision (ICCV)},
    month     = {October},
    year      = {2025},
    pages     = {24655-24665}
}

@ARTICLE{Xu_beamspace_ddma2021,
  author={Xu, Feng and Vorobyov, Sergiy A. and Yang, Fawei},
  journal={IEEE Transactions on Vehicular Technology}, 
  title={Transmit Beamspace DDMA Based Automotive MIMO Radar}, 
  year={2022},
  volume={71},
  number={2},
  pages={1669-1684},
  doi={10.1109/TVT.2021.3132886}
}

@inproceedings{kradar2022,
    author = {Paek, Dong-Hee and KONG, SEUNG-HYUN and Wijaya, Kevin Tirta},
    booktitle = {Advances in Neural Information Processing Systems},
    editor = {S. Koyejo and S. Mohamed and A. Agarwal and D. Belgrave and K. Cho and A. Oh},
    pages = {3819--3829},
    publisher = {Curran Associates, Inc.},
    title = {K-Radar: 4D Radar Object Detection for Autonomous Driving in Various Weather Conditions},
    volume = {35},
    year = {2022}
}

@ARTICLE{radelft2024,
    author={Roldan, Ignacio and Palffy, Andras and Kooij, Julian F. P. and Gavrila, Dariu M. and Fioranelli, Francesco and Yarovoy, Alexander},
    journal={IEEE Transactions on Radar Systems}, 
    title={A Deep Automotive Radar Detector Using the RaDelft Dataset}, 
    year={2024},
    volume={2},
    number={},
    pages={1062-1075},
    doi={10.1109/TRS.2024.3485578}
}

@InProceedings{pointnet2017,
    author = {Qi, Charles R. and Su, Hao and Mo, Kaichun and Guibas, Leonidas J.},
    title = {PointNet: Deep Learning on Point Sets for 3D Classification and Segmentation},
    booktitle = {Proceedings of the IEEE Conference on Computer Vision and Pattern Recognition (CVPR)},
    month = {July},
    year = {2017}
}

@INPROCEEDINGS{rpfa_net2021,
    author={Xu, Baowei and Zhang, Xinyu and Wang, Li and Hu, Xiaomei and Li, Zhiwei and Pan, Shuyue and Li, Jun and Deng, Yongqiang},
    booktitle={2021 IEEE International Intelligent Transportation Systems Conference (ITSC)}, 
    title={RPFA-Net: a 4D RaDAR Pillar Feature Attention Network for 3D Object Detection}, 
    year={2021},
    volume={},
    number={},
    pages={3061-3066},
    doi={10.1109/ITSC48978.2021.9564754}
}

@INPROCEEDINGS{radarpillars2024,
  author={Musiat, Alexander and Reichardt, Laurenz and Schulze, Michael and Wasenmüller, Oliver},
  booktitle={2024 IEEE 27th International Conference on Intelligent Transportation Systems (ITSC)}, 
  title={RadarPillars: Efficient Object Detection from 4D Radar Point Clouds}, 
  year={2024},
  volume={},
  number={},
  pages={1656-1663},
  doi={10.1109/ITSC58415.2024.10919920}
}

@article{kde_chen2017,
  title={A tutorial on kernel density estimation and recent advances},
  author={Chen, Yen-Chi},
  journal={Biostatistics \& Epidemiology},
  volume={1},
  number={1},
  pages={161--187},
  year={2017},
  publisher={Taylor \& Francis}
}

@INPROCEEDINGS{cruw3d2024,
    author={Wang, Yizhou and Cheng, Jen-Hao and Huang, Jui-Te and Kuan, Sheng-Yao and Fu, Qiqian and Ni, Chiming and Hao, Shengyu and Wang, Gaoang and Xing, Guanbin and Liu, Hui and Hwang, Jenq-Neng},
    booktitle={2024 IEEE Intelligent Vehicles Symposium (IV)}, 
    title={Vision meets mmWave Radar: 3D Object Perception Benchmark for Autonomous Driving}, 
    year={2024},
    volume={},
    number={},
    pages={2769-2775},
    doi={10.1109/IV55156.2024.10588620}
}

@INPROCEEDINGS{scorp2020,
    author={Nowruzi, Farzan Erlik and Kolhatkar, Dhanvin and Kapoor, Prince and Al Hassanat, Fahed and Heravi, Elnaz Jahani and Laganiere, Robert and Rebut, Julien and Malik, Waqas},
    booktitle={2020 IEEE MTT-S International Conference on Microwaves for Intelligent Mobility (ICMIM)}, 
    title={Deep Open Space Segmentation using Automotive Radar}, 
    year={2020},
    volume={},
    number={},
    pages={1-4},
    doi={10.1109/ICMIM48759.2020.9299052}
}

@INPROCEEDINGS{rodnet2021,
    author={Wang, Yizhou and Jiang, Zhongyu and Gao, Xiangyu and Hwang, Jenq-Neng and Xing, Guanbin and Liu, Hui},
    booktitle={2021 IEEE Winter Conference on Applications of Computer Vision (WACV)}, 
    title={RODNet: Radar Object Detection using Cross-Modal Supervision}, 
    year={2021},
    volume={},
    number={},
    pages={504-513},
    doi={10.1109/WACV48630.2021.00055}
}

@InProceedings{Dong2020,
    author = {Dong, Xu and Wang, Pengluo and Zhang, Pengyue and Liu, Langechuan},
    title = {Probabilistic Oriented Object Detection in Automotive Radar},
    booktitle = {Proceedings of the IEEE/CVF Conference on Computer Vision and Pattern Recognition (CVPR) Workshops},
    month = {June},
    year = {2020}
}

@inproceedings{madani2022radatron,
  title={Radatron: Accurate detection using multi-resolution cascaded MIMO radar},
  author={Madani, Sohrab and Guan, Jayden and Ahmed, Waleed and Gupta, Saurabh and Hassanieh, Haitham},
  booktitle={European Conference on Computer Vision},
  pages={160--178},
  year={2022},
  organization={Springer}
}

@ARTICLE{transrad2025,
  author={Cheng, Lei and Cao, Siyang},
  journal={IEEE Transactions on Radar Systems}, 
  title={TransRAD: Retentive Vision Transformer for Enhanced Radar Object Detection}, 
  year={2025},
  volume={3},
  number={},
  pages={303-317},
  doi={10.1109/TRS.2025.3537604}
}

@InProceedings{transradar2024,
    author    = {Dalbah, Yahia and Lahoud, Jean and Cholakkal, Hisham},
    title     = {TransRadar: Adaptive-Directional Transformer for Real-Time Multi-View Radar Semantic Segmentation},
    booktitle = {Proceedings of the IEEE/CVF Winter Conference on Applications of Computer Vision (WACV)},
    month     = {January},
    year      = {2024},
    pages     = {353-362}
}

@INPROCEEDINGS{mvrae2024,
  author={Zhu, Haoran and He, Haoze and Choromanska, Anna and Ravindran, Satish and Shi, Binbin and Chen, Lihui},
  booktitle={2024 IEEE Intelligent Vehicles Symposium (IV)}, 
  title={Multi-View Radar Autoencoder for Self-Supervised Automotive Radar Representation Learning}, 
  year={2024},
  volume={},
  number={},
  pages={1601-1608},
  doi={10.1109/IV55156.2024.10588463}
}

@InProceedings{dps2021,
    title = 	 {Active Deep Probabilistic Subsampling},
    author =       {Van Gorp, Hans and Huijben, Iris and Veeling, Bastiaan S and Pezzotti, Nicola and Van Sloun, Ruud J. G.},
    booktitle = 	 {Proceedings of the 38th International Conference on Machine Learning},
    pages = 	 {10509--10518},
    year = 	 {2021},
    editor = 	 {Meila, Marina and Zhang, Tong},
    volume = 	 {139},
    series = 	 {Proceedings of Machine Learning Research},
    month = 	 {18--24 Jul},
    publisher =    {PMLR}
}

@article{hendrycks2016gaussian,
  title={Gaussian error linear units (gelus)},
  author={Hendrycks, Dan and Gimpel, Kevin},
  journal={arXiv preprint arXiv:1606.08415},
  year={2016}
}

@inproceedings{AdamW_loschilov2019,
    title={Decoupled Weight Decay Regularization},
    author={Ilya Loshchilov and Frank Hutter},
    booktitle={International Conference on Learning Representations},
    year={2019}
}

@inproceedings{onecyclelr_smith2019,
  title={Super-convergence: Very fast training of neural networks using large learning rates},
  author={Smith, Leslie N and Topin, Nicholay},
  booktitle={Artificial intelligence and machine learning for multi-domain operations applications},
  volume={11006},
  pages={369--386},
  year={2019},
  organization={SPIE}
}

@InProceedings{t_fftradnet2023,
    author    = {Giroux, James and Bouchard, Martin and Laganiere, Robert},
    title     = {T-FFTRadNet: Object Detection with Swin Vision Transformers from Raw ADC Radar Signals},
    booktitle = {Proceedings of the IEEE/CVF International Conference on Computer Vision (ICCV) Workshops},
    month     = {October},
    year      = {2023},
    pages     = {4030-4039}
}
}


\end{document}